\documentclass{article}

\usepackage{microtype}
\usepackage{graphicx}
\usepackage{subcaption}
\usepackage{booktabs}

\usepackage{hyperref}

\usepackage[accepted]{icml2026}

\usepackage{amsmath}
\usepackage{amssymb}
\usepackage{mathtools}
\usepackage{amsthm}

\usepackage[capitalize,noabbrev]{cleveref}

\theoremstyle{plain}

\theoremstyle{definition}

\theoremstyle{remark}

\icmltitlerunning{Weight-Space Geometry of Offline Reasoning Training}

\begin{document}

\twocolumn[
  \icmltitle{Weight-Space Geometry of Offline Reasoning Training}

  \icmlsetsymbol{equal}{*}

  \begin{icmlauthorlist}
    \icmlauthor{Aleksandr Nikolich}{keen}
    \icmlauthor{Igor Kiselev}{acc}
    \icmlauthor{Vladimir Platonov}{yndx}
    \icmlauthor{Karina Romanova}{yndx}
  \end{icmlauthorlist}

  \icmlaffiliation{keen}{Keenable AI}
  \icmlaffiliation{acc}{Accenture}
  \icmlaffiliation{yndx}{Yandex}

  \icmlcorrespondingauthor{Aleksandr Nikolich}{alexn@keenable.ai}

  \icmlkeywords{Mechanistic Interpretability, Offline RL, Reasoning, LoRA, Weight Space}

  \vskip 0.3in
]

\printAffiliationsAndNotice{}

\begin{abstract}
  Offline reinforcement-learning losses (RFT, RIFT, DFT, Offline GRPO, DPO) are widely used to distill reasoning from large teachers into smaller students, and are typically compared on downstream accuracy alone. We ask whether they are mechanistically distinct or converge to a similar weight update. Training six methods (SFT, RFT, DFT, RIFT, Offline GRPO, DPO) on identical math rollouts from a single base model (Qwen3-4B) with attention-only LoRA, we analyze the resulting deltas via cosine similarity, principal-angle subspace analysis, linear mode connectivity, and CKA. We observe: (i) SFT, RFT, and RIFT have nearly colinear weight deltas (cosine $\geq 0.97$, top-$1$ principal angle ${\sim}\!\!7^\circ$ median over $144$ modules) and comparable GSM8K accuracy ($87$--$88\%$, $n{=}1319$; pairwise McNemar $p \geq 0.15$); (ii) DFT diverges further in direction than any reward-weighted method despite using the same data; (iii) Offline GRPO adds a substantial component orthogonal to the SFT direction ($\sim\!\!67\%$ globally, up to $\sim\!\!86\%$ in late layers) while staying in the SFT loss basin; (iv) DPO sits in a near-orthogonal subspace, shows a mode-connectivity barrier, and collapses late-layer CKA to $\sim 0.46$. DPO also reaches the highest accuracy in our protocol on both GSM8K ($93.5\%$, McNemar $p < 10^{-9}$ vs.\ each other method) and AIME26 ($30.0\%$ vs.\ $3.3$--$10.0\%$); its training uses a $10\times$ smaller learning rate than the others (the standard convention), so the update-norm and accuracy gaps reflect loss-function and optimizer choices jointly, and a learning-rate-matched DPO comparison is left for future work.
\end{abstract}

\section{Introduction}
\label{sec:intro}

Reasoning distillation has become a standard recipe for teaching small models to solve math and code tasks: a strong teacher generates rollouts, and a student is trained on them with one of a rapidly growing list of offline objectives. The past year alone introduced RIFT~\citep{rift2026}, Offline GRPO~\citep{krafton-grpo}, DFT~\citep{dft2025}, LUFFY~\citep{luffy2025}, and DAPO~\citep{dapo2025}, alongside an established preference-learning family --- DPO~\citep{dpo2023}, KTO~\citep{kto2024}, IPO~\citep{ipo2023}, and NCA~\citep{infonca2024} --- each accompanied by claims that its specific loss formulation is responsible for accuracy gains over plain SFT.

These methods are compared almost exclusively by benchmark accuracy. What they do to the model is unknown: do different losses produce weight updates that point in the same direction, or qualitatively different ones? The distinction matters for both practitioners (which loss is worth implementing?) and interpretability researchers (does ``offline RL'' name a single mechanism or a family?).

We present a controlled weight-space comparison of offline reasoning losses: identical rollouts, identical base model (Qwen3-4B-Instruct), shared LoRA initialization, six methods (DPO uses a smaller learning rate per its codebase convention, see \S\ref{sec:setup}). Following recent weight-space studies of fine-tuning~\citep{sharedsubspaces2025,convergent-misalignment2025,watchweights2025,rank1lora2025}, we analyze each method's LoRA delta $\Delta W$ rather than its outputs.

Our contributions are: (1) reward-weighted losses (SFT, RFT, RIFT) converge on essentially the same direction in weight space (cosine $\geq 0.97$) and produce GSM8K accuracies that are non-different by exact McNemar's test ($p \geq 0.15$, $n{=}1319$); (2) DFT, despite being a one-line modification of SFT, produces a more distinctive update than any explicitly reward-weighted method; (3) Offline GRPO adds a quantifiable orthogonal component (globally $67\%$, rising to $\sim\!\!80\%$ in late layers) while staying in the same loss basin as SFT/RIFT; (4) DPO sits in a near-orthogonal subspace with higher effective rank, a sharp linear-mode barrier, and reaches the highest pass@1 on both GSM8K and AIME26 in our protocol; we report this with the caveat that DPO uses a $10\times$ smaller learning rate, so the loss formulation and optimizer setting cannot be cleanly separated here.

\section{Setup}
\label{sec:setup}

\paragraph{Data.} All methods share one set of rollouts: DeepScaleR prompts (${\sim}40$k verified math~\citep{deepscaler2025}), teacher DeepSeek-V4-Flash, $K{=}4$ CoT completions/prompt, binary \texttt{math-verify} reward. Reference-policy methods use $\pi_{\text{base}}$. DPO consumes ${\sim}1.8$K (chosen, rejected) pairs vs.\ ${\sim}75$K rows for the rest. Identical rollouts is the central control.

\paragraph{Methods.} Table~\ref{tab:methods} summarizes the six losses. With $\ell_i=-\log\pi_\theta(y_i\mid x)$ shorthand for the per-sequence NLL: SFT $=\sum_i \ell_i$ on all $i$; RFT~\citep{rft2023} $=\sum_{i:r_i=1}\ell_i$ on positives only; DFT~\citep{dft2025} $=\sum_t \mathrm{sg}(\pi_\theta(y_t\mid y_{<t},x))\,\ell_t^{\text{tok}}$ down-weights confident tokens; RIFT~\citep{rift2026} $=\sum_i (1-r_i\,\lambda)\,\ell_i$ is a linear-in-reward surrogate that admits negatives; Offline GRPO~\citep{krafton-grpo,deepseekmath} $=\sum_i \hat A_i\,\ell_i$ with $\hat A_i = r_i - \bar r_g + b$; DPO~\citep{dpo2023} $=-\log\sigma\!\left(\beta\,[\log\frac{\pi_\theta(y_w)}{\pi_{\text{ref}}(y_w)} - \log\frac{\pi_\theta(y_l)}{\pi_{\text{ref}}(y_l)}]\right)$ on (chosen $y_w$, rejected $y_l$) pairs.

\begin{table}[t]
  \caption{Offline reasoning losses studied. ``Neg.'' = uses negative samples; ``Rew.'' = uses scalar reward; ``Ref.'' = needs a reference policy.}
  \label{tab:methods}
  \vskip 0.05in
  \centering
  \footnotesize
  \begin{tabular}{lcccp{2.4cm}}
    \toprule
    Method & Neg. & Rew. & Ref. & Key idea \\
    \midrule
    SFT          & --        & --   & --   & MLE on all rollouts \\
    RFT          & filter    & impl.& --   & MLE on positives \\
    DFT          & down-w.   & --   & --   & $\mathcal L \!\cdot\! \mathrm{sg}(\pi_\theta)$ \\
    RIFT         & weight    & yes  & --   & Reward-weighted MLE \\
    Off.\ GRPO   & yes       & yes  & yes  & Group-relative adv. \\
    DPO          & paired    & impl.& yes  & Contrastive log-ratio \\
    \bottomrule
  \end{tabular}
  \vskip -0.1in
\end{table}

\paragraph{Training.} Qwen3-4B-Instruct-2507, LoRA on attention projections (\texttt{q,k,v,o\_proj}; rank $32$, $\alpha{=}64$, dropout $0$; $144$ modules over $36$ layers). Effective batch $32$, cosine schedule, $5\%$ warmup, wd $0.01$, grad-clip $1.0$, seed $42$, bf16. Peak LR $5\!\times\!10^{-6}$ for all but DPO ($5\!\times\!10^{-7}$, codebase convention; higher diverges). DPO: sigmoid loss, $\beta{=}0.1$; Offline GRPO: additive bias $0.1$ on the centered advantage, no probability weighting, no explicit KL. $1{,}500$ steps; we report step $1{,}000$ uniformly.

\paragraph{Analysis.} Let $\Delta W^{(m)}=B^{(m)}A^{(m)}$ be the stacked LoRA delta of method $m$. We measure: \textbf{(i)}~global/per-layer cosine $\langle \Delta W^{(m)}, \Delta W^{(m')}\rangle / \|\Delta W^{(m)}\|\|\Delta W^{(m')}\|$; \textbf{(ii)}~per-layer SVD (effective rank, principal angles between top-$k$ subspaces); \textbf{(iii)}~linear mode connectivity~\citep{frankle2020lmc}: masked-answer CE on GSM8K along $\alpha \Delta W^{(m)} + (1{-}\alpha)\Delta W^{(m')}$; \textbf{(iv)}~CKA~\citep{kornblith2019cka} of merged-model hidden states.

\section{Results}
\label{sec:results}

\subsection{Downstream accuracy}
Figure~\ref{fig:accuracy} (appendix) reports greedy pass@1 on full GSM8K ($n{=}1319$) and AIME26 ($n{=}30$). SFT/RFT/DFT/RIFT/Offline GRPO sit at $87.3$--$88.2\%$ on GSM8K, pairwise non-different by exact McNemar ($p \geq 0.15$); DPO reaches $93.5\%$ ($p < 10^{-9}$). On AIME26 the ordering repeats but $n{=}30$ is underpowered (SFT--DPO $p{=}0.07$). DPO trains at a $10\times$ smaller LR with ${\sim}40\times$ fewer rows, so we treat the gap as suggestive. Llama-3.2-3B replicates the geometry and the $5$--$7$ point DPO accuracy edge.

\paragraph{On-policy RL preserves accuracy; SFT-style loses it.} Re-measuring greedy pass@1 for the on-policy methods (Table~\ref{tab:online-acc}, our adapters, same protocol) shows their reward-orthogonal updates (\S\ref{sec:seed-lr}) do \emph{not} cost accuracy: Online GRPO/DAPO and DPO all stay at the base instruct model's $93$--$94\%$ on GSM8K, whereas SFT and Offline GRPO drop to ${\sim}87\%$ (below base). Online GRPO is best on AIME26 ($20.0\%$).

\begin{table}[h]
\small\centering
\vskip -0.02in
\begin{tabular}{lcc}
\toprule
Method & GSM8K & AIME26 \\
\midrule
Base instruct & 94.0 & 16.7 \\
SFT \,/\, Offline GRPO & 87.6 / 87.3 & 6.7 / 6.7 \\
DPO & 94.2 & 13.3 \\
Online GRPO & 93.7 & \textbf{20.0} \\
Online DAPO & 93.3 & 16.7 \\
\bottomrule
\end{tabular}
\caption{Greedy pass@1 (\%) for our consistently-trained adapters; GSM8K $n{=}1319$, AIME26 $n{=}30$. SFT-direction methods sit below the base model; reward-orthogonal methods match or beat it.}
\label{tab:online-acc}
\end{table}

\subsection{Weight-space convergence}
\label{sec:weightspace}
\label{sec:cosine}

\begin{figure}[t]
  \centering
  \includegraphics[width=0.92\columnwidth]{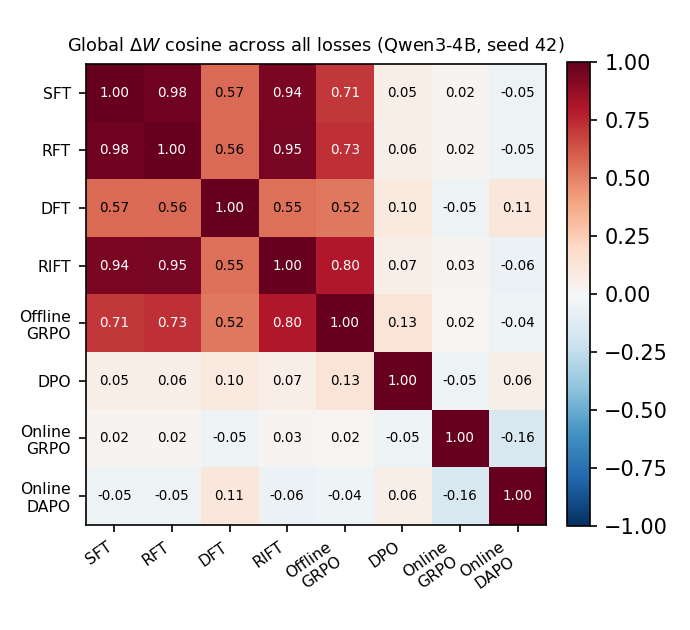}
  \vskip -0.04in
  \caption{Global $\Delta W$ cosine across all eight losses (Qwen3-4B, seed $42$; all adapters trained in one consistent space). Reward-weighted SFT/RFT/RIFT cluster ($0.94$--$0.98$); DFT intermediate ($\sim\!0.55$); \emph{Offline} GRPO at $0.71$--$0.80$ to the cluster; DPO near-orthogonal ($\leq 0.13$). The two \emph{on-policy} methods, \emph{Online} GRPO and \emph{Online} DAPO, are each near-orthogonal to every offline loss \emph{and to each other} ($-0.16$); orthogonal-fraction off SFT is $0.69$ (offline GRPO) vs.\ $0.998$/$0.995$ (online GRPO/DAPO). On-policy sampling, not the group-relative loss, drives the departure from SFT.}
  \label{fig:cosine_global}
  \label{fig:online-grpo}
  \vskip -0.05in
\end{figure}

Figure~\ref{fig:cosine_global} shows the cosine similarity matrix between global LoRA deltas. Three regimes are visible. \emph{First}, SFT, RFT, and RIFT form a tight cluster: the SFT--RFT, SFT--RIFT, and RFT--RIFT cosines are $0.977$, $0.967$, $0.969$. Filtering negatives (RFT) and reward-weighting them (RIFT) does not measurably change the direction of the update relative to plain SFT on the union; it only adjusts the step size, with $\|\Delta W\|_F$ ranging from $2.82$ (RFT) to $3.04$ (RIFT). \emph{Second}, DFT, which differs from SFT by a single multiplicative factor on the loss, sits at cosine $0.572$ to SFT and $0.536$ to RIFT --- a larger directional change than any explicitly reward-weighted method. \emph{Third}, DPO is orthogonal to everything: cosine to SFT, RFT, and RIFT all fall in $[0.057, 0.065]$. Offline GRPO occupies an intermediate position (cosine $\approx 0.74$ to the SFT cluster).

The per-layer view (Figure~\ref{fig:cosine_per_layer}) decomposes this further. SFT/RFT/RIFT pairs are essentially flat above $0.95$ at every layer. SFT--GRPO drops gradually with depth. SFT--DFT is bimodal --- close to $1$ at certain bottleneck layers and well below $0.5$ at others. SFT--DPO hovers near zero throughout. The visible drop on layers $34$--$36$ across all pairs reflects the small-norm tail of LoRA updates in the last decoder block; we treat this as a LoRA artifact rather than a finding.

\subsection{Subspace analysis}
\label{sec:subspace}

Per-layer SVD lets us go beyond a single direction and ask whether two methods adapt the same low-dimensional subspace. We report principal angles between the top-$10$ left singular vectors of each $\Delta W^{(m)}$ at the same module. Smaller angles mean shared subspace.

Aggregating across all $144$ attention modules (top-$10$ left singular vectors per module), median top-$1$ principal angles are $6.7^\circ$ (SFT--RFT), $8.2^\circ$ (SFT--RIFT), $18.5^\circ$ (SFT--Offline GRPO), $26.7^\circ$ (SFT--DFT), and $54.6^\circ$ (SFT--DPO); the median worst (top-$10$) angles are $36^\circ$, $40^\circ$, $76^\circ$, $85^\circ$, $90^\circ$ in the same order. SFT--DPO IQR for the worst angle is $[89.6^\circ, 89.8^\circ]$: essentially every module is orthogonal at every singular index. The reward-weighted cluster shares the top of its subspace with SFT to within ${\sim}\!\!10^\circ$; GRPO and DFT partially overlap; DPO does not.

The effective rank, averaged over all $144$ modules, is $\sim 16$ for SFT, RFT, DFT, and RIFT, $14.8$ for Offline GRPO, and $24.5$ for DPO. DPO writes into a higher-dimensional subspace, but, given its $13\times$ smaller Frobenius norm, with much smaller singular values; together with the orthogonality to SFT, this suggests DPO learns a different decomposition of the same projection matrices rather than a low-rank refinement of SFT.

To quantify how much of Offline GRPO's update is genuinely new direction, we project $\Delta W^{\mathrm{grpo}}$ onto the SFT direction at every adapted module and report $\|\Delta W^{\mathrm{grpo}} - \Pi_{\mathrm{sft}}\Delta W^{\mathrm{grpo}}\|_F / \|\Delta W^{\mathrm{grpo}}\|_F$. Globally this is $0.67$; per layer it grows from $\sim\!\!0.55$ in middle blocks to $0.79$--$0.86$ in the final five blocks --- the same layers where CKA diverges (Section~\ref{sec:cka}).

\paragraph{Top-1 singular directions.} The rank-1 approximation $\Delta W \approx \sigma_1 u_1 v_1^\top$ isolates the single most important output direction $u_1$ each loss writes into. Mean $|\langle u_1^m, u_1^{m'}\rangle|$ over $144$ modules is $0.97$--$0.98$ within SFT/RFT/RIFT, $0.78$--$0.80$ to Offline GRPO, $0.64$--$0.67$ to DFT, and $0.11$ to DPO. Right singular vectors $v_1$ (input directions) converge much more tightly: $0.99$--$1.00$ for non-DPO pairs, $0.94$ for DFT, $0.66$--$0.70$ for DPO. All methods (except DPO) read from nearly the same input subspace; they differ in how they transform it.

\subsection{Seed and learning-rate sensitivity}
\label{sec:seed-lr}

The colinearity above is at a \emph{single} seed, conflating loss agreement with shared-init
agreement. We disentangle by training each loss at two seeds ($42,123$) and three LRs
($5\!\times\!10^{-7..-5}$); $\Delta W=(\alpha/r)BA$ is gauge-invariant, so its cosine is genuine.

\begin{figure*}[t]
  \centering
  \includegraphics[width=0.30\textwidth]{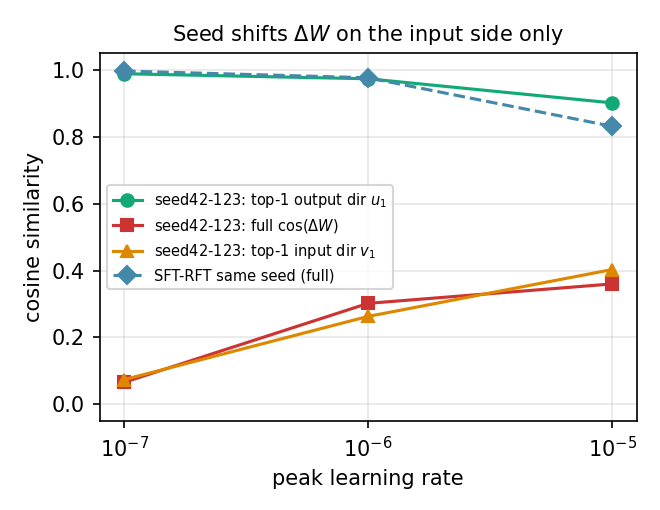}\hfill
  \includegraphics[width=0.30\textwidth]{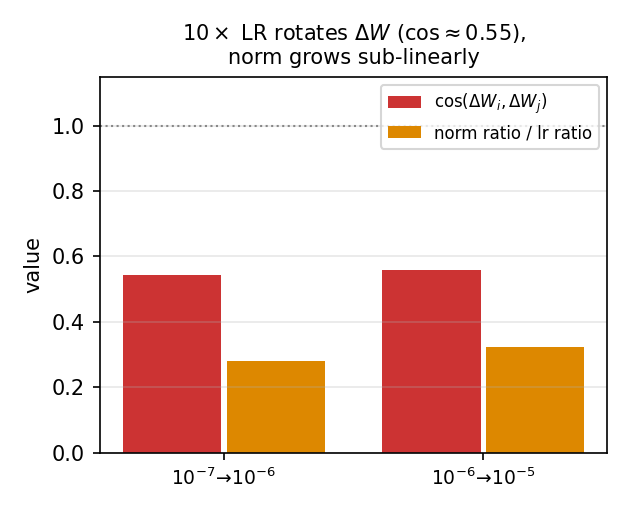}\hfill
  \includegraphics[width=0.30\textwidth]{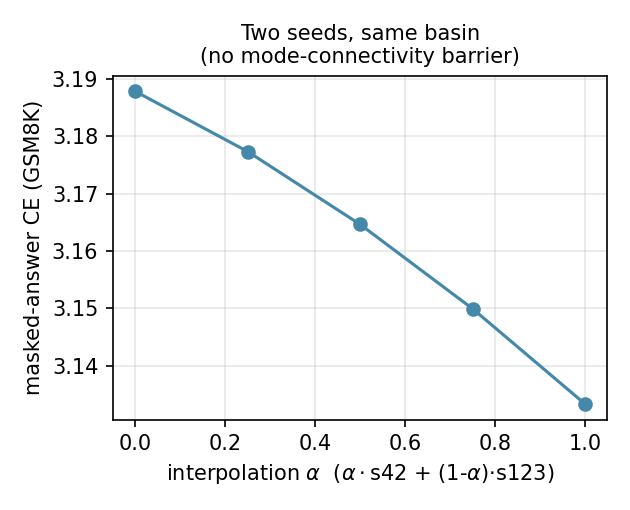}
  \caption{\textbf{Seed and learning-rate sensitivity (SFT, Qwen3-4B).} \emph{Left:} across two
  seeds the output direction $u_1$ stays aligned ($\sim\!0.99$) while the input direction $v_1$
  and full cosine are low at small LR and rise with LR; dashed shows SFT--RFT at a fixed seed.
  \emph{Middle:} a $10\times$ LR step rotates $\Delta W$ (cosine $\approx 0.55$) and grows its
  norm sub-linearly --- LR is not a pure rescaling. \emph{Right:} interpolating the two seeds'
  deltas shows no loss barrier --- different weights, same basin.}
  \label{fig:seed-lr}
\end{figure*}

\textbf{Seed rotates $\Delta W$ more than the loss --- but only on the input side.} At a fixed
seed SFT--RFT are colinear (cosine $0.996$, angle $3.7^\circ$), yet the \emph{same loss at two
seeds} has cosine only $0.07$ ($5\!\times\!10^{-7}$)--$0.36$ ($5\!\times\!10^{-5}$). Cause: LoRA's
random $A$ init --- across seeds the top-$1$ \emph{output} direction $u_1$ still agrees at $0.99$
while the \emph{input} direction $v_1$ agrees at $0.07$ (median top-$8$ angle $26^\circ$ vs.\
$76^\circ$ for unrelated runs). Functionally the seeds are the \emph{same} solution: interpolating
their deltas shows no barrier (midpoint $+0.004$). So the cross-method colinearity is partly
shared-init, but convergence onto a common output subspace is seed-robust (Figure~\ref{fig:seed-lr}).

\textbf{Learning rate changes direction, not just magnitude.} A $10\times$ LR step rotates
$\Delta W$ (cosine $\approx 0.55$) and grows its norm only $\sim\!3\times$ --- not a pure
rescaling, which sharpens the caveat on the $10\times$-smaller-LR DPO comparison.

\textbf{Online GRPO is far more orthogonal than offline GRPO.} We also train \emph{online}
GRPO under the same LoRA recipe (on-policy rollouts, group-relative advantage,
\texttt{math\_verify} reward; $600$ steps, $8$ generations/prompt, lr $5\!\times\!10^{-6}$, seed
$42$) --- the comparison the original protocol could not produce. The resulting update is almost
entirely orthogonal to the SFT/RFT cluster: cosine $0.025$ to SFT and $0.024$ to RFT, with an
orthogonal fraction of $0.998$ off the SFT direction (Figure~\ref{fig:online-grpo}), versus
$0.67$ for \emph{offline} GRPO (\S\ref{sec:weightspace}). Its Frobenius norm is $\sim\!10\times$
smaller than SFT's at the same LR ($0.30$ vs.\ $2.84$), echoing the small-norm regime of DPO.
On-policy sampling thus moves the update off the shared SFT subspace far more than the offline
group-relative loss does, indicating that the SFT/offline-RL directional convergence is partly a
consequence of training on the \emph{same fixed rollouts}: replacing them with on-policy samples
largely breaks it.

\subsection{Linear mode connectivity}
\label{sec:lmc}

We linearly interpolate LoRA deltas, merge into the base, and measure the per-token cross-entropy of the gold $\backslash\mathtt{boxed\{answer\}}$ continuation right after the prompt on GSM8K. The metric is length-sensitive: DPO produces longer, structured CoTs (median $5100$ vs $1100$ chars on correct AIME26; verification steps in $9/9$ correct DPO solutions vs $0/3$ SFT), inflating per-token NLL of the bare boxed answer. Offline GRPO $\to$ RIFT improves monotonically ($4.93 \to 2.25$) and SFT $\to$ Offline GRPO worsens monotonically ($2.06 \to 4.93$): one basin. RIFT $\to$ DPO shows a sharp non-monotonic barrier above $\alpha{=}0.5$ ($3.82 \to 7.06 \to 8.64$): even discounting length, linear interpolation destroys the solution.

\subsection{Representational similarity}
\label{sec:cka}

CKA on hidden states (Figure~\ref{fig:cka}, $100$ GSM8K prompts) confirms the weight-space picture. SFT--RIFT CKA stays above $0.99$ at every layer; SFT/RIFT--Offline GRPO drop to $\sim 0.85$ in the final layer (GRPO reshapes output-facing layers); Offline GRPO--DPO and DFT--DPO start near $0.93$ and collapse to $\sim\!\!0.45$ from layer $25$ onward. A logit-lens probe gives a complementary null --- mean prediction depth is $35.3$--$36.0$ out of $36$ for every method --- but this is partly forced by attention-only LoRA leaving MLPs frozen, so it should not be read as a finding about the losses.

\section{Discussion}
\label{sec:discussion}

\paragraph{Reward-weighted MLE is SFT, plus DFT is the surprise.} SFT, RFT, and RIFT differ only in how they handle negative samples (drop, weight, or include uniformly), yet the resulting LoRA deltas have cosine $\geq 0.97$, principal angles $<25^\circ$, and indistinguishable per-layer CKA, and they sit within $1$ percentage point of each other on full GSM8K. The Frobenius norms differ by up to $7\%$. If RIFT outperforms SFT at the same step count, our results suggest the explanation lies in magnitude (effective step size in the SFT direction), not direction --- a longer or higher-lr SFT run should close the gap. DFT, by contrast, has cosine $\sim 0.55$ to SFT/RIFT despite using \emph{less} information than they do (no reward, no filtering): self-weighting by $\mathrm{sg}(\pi_\theta)$ reweights \emph{which} examples drive the update in a way explicit reward does not, yet leaves the loss basin unchanged.

\paragraph{Offline GRPO shifts direction but stays in basin; online GRPO does not.} Among offline rewards, only Offline GRPO substantially shifts direction from SFT (cosine $0.73$; orthogonal-fraction $0.67$, $\sim\!\!0.8$ late; angles up to $59^\circ$), yet barrier-free interpolations keep it in the SFT/RIFT basin. \emph{Online} GRPO goes much further --- orthogonal fraction $0.998$ (Figure~\ref{fig:online-grpo}) --- so the SFT/offline-RL convergence is partly an artifact of shared fixed rollouts, which on-policy sampling breaks.

\paragraph{DPO sits apart, geometrically and on accuracy.} DPO occupies a near-orthogonal subspace ($74^\circ$--$89^\circ$), a higher-rank update with much smaller Frobenius norm, a sharp linear-mode barrier, and late-layer CKA $\sim 0.45$; it also reaches the highest pass@1 on GSM8K ($93.5\%$) and AIME26 ($30.0\%$). It trains at a $10\times$ smaller LR, so its norm/accuracy gaps are entangled with the optimizer --- correlation worth a LR-matched follow-up, not a causal claim. Extending the weight-space view to the wider contrastive family (IPO, KTO, SimPO, Cal-DPO) is left open.

\paragraph{Limitations.} Single domain and checkpoint; attention-only LoRA; greedy-only on small AIME26 ($n{=}30$). DPO uses $10\times$ smaller LR and ${\sim}40\times$ fewer rows, so its norm/accuracy gaps are entangled with the optimizer. Online GRPO is reported at lr $5\!\times\!10^{-6}$, seed $42$ (\S\ref{sec:seed-lr}); the remaining LR/seed cells, a matched accuracy comparison, and calibrated DPO variants (KTO, IPO, SimPO) are left to a fuller sweep. Code, adapters, and analysis scripts are released.

\bibliographystyle{icml2026}
\bibliography{references}

\appendix
\section{Supplementary figures}
\label{app:figs}

\begin{figure}[h]
  \centering
  \includegraphics[width=\linewidth]{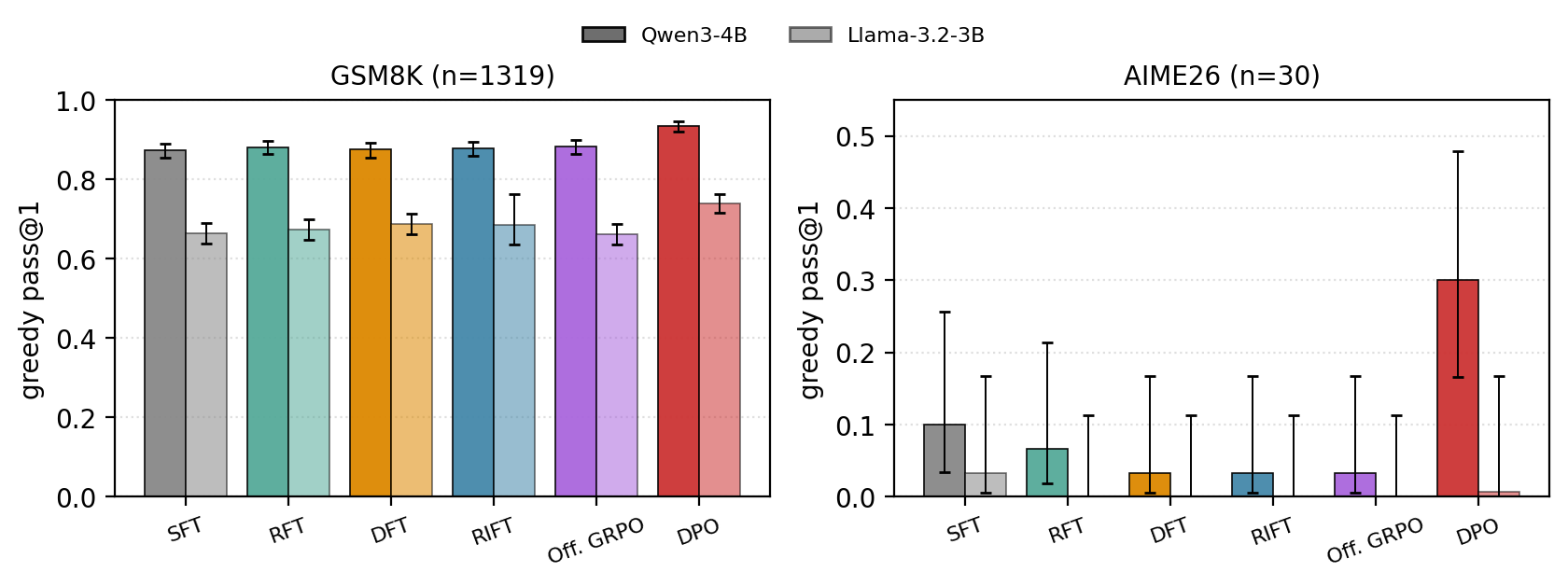}
  \caption{Greedy pass@1 with Wilson $95\%$ CI bars on GSM8K ($n{=}1319$) and AIME26 ($n{=}30$). Dark bars: Qwen3-4B-Instruct. Light bars: Llama-3.2-3B-Instruct. On both architectures, DPO sits noticeably above the SFT/RFT/DFT/RIFT/Offline GRPO cluster on GSM8K (Qwen3: McNemar $p<10^{-9}$ vs.\ each other method); Llama-3.2-3B AIME26 floors near zero at this model scale.}
  \label{fig:accuracy}
\end{figure}

\begin{figure}[h]
  \centering
  \includegraphics[width=\linewidth]{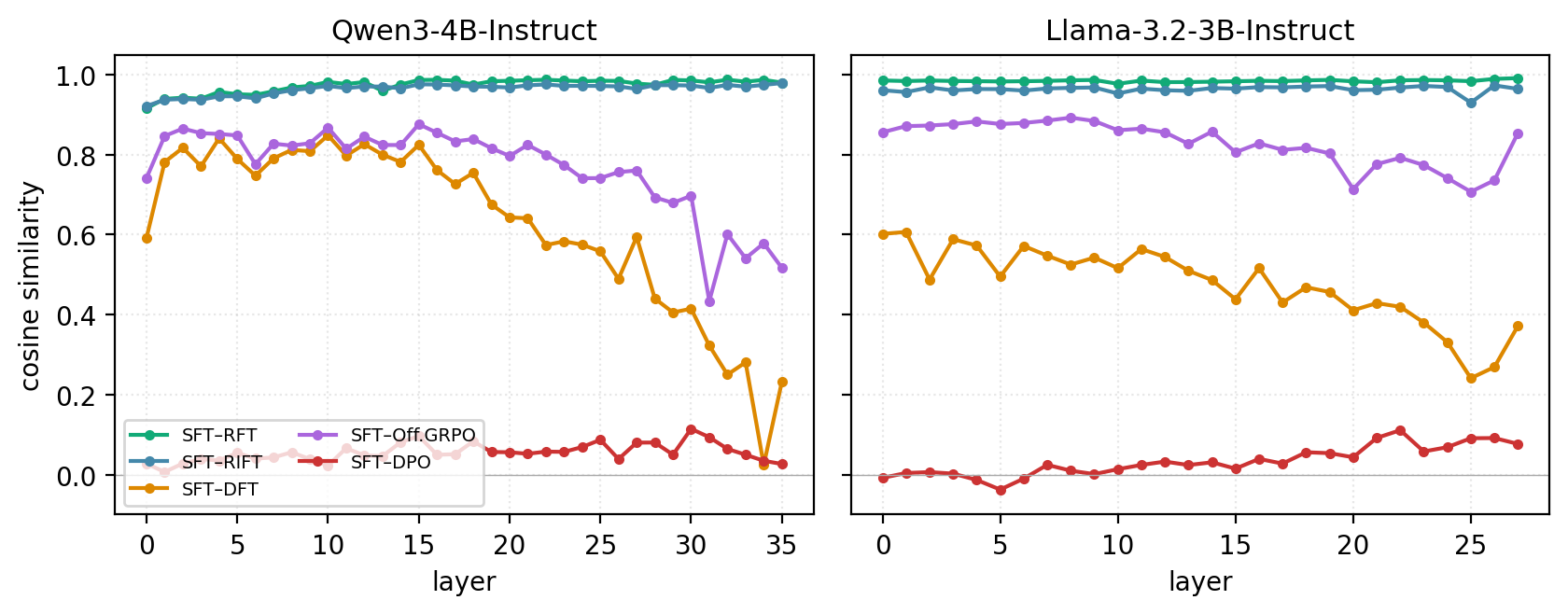}
  \caption{Per-layer cosine similarity of LoRA deltas to SFT, on Qwen3-4B (left, $36$ layers) and Llama-3.2-3B (right, $28$ layers). SFT/RFT/RIFT track each other across all layers; Offline GRPO, DFT, and especially DPO diverge in deeper layers, with the same qualitative pattern on both architectures.}
  \label{fig:cosine_per_layer}
\end{figure}

\begin{figure}[h]
  \centering
  \includegraphics[width=\linewidth]{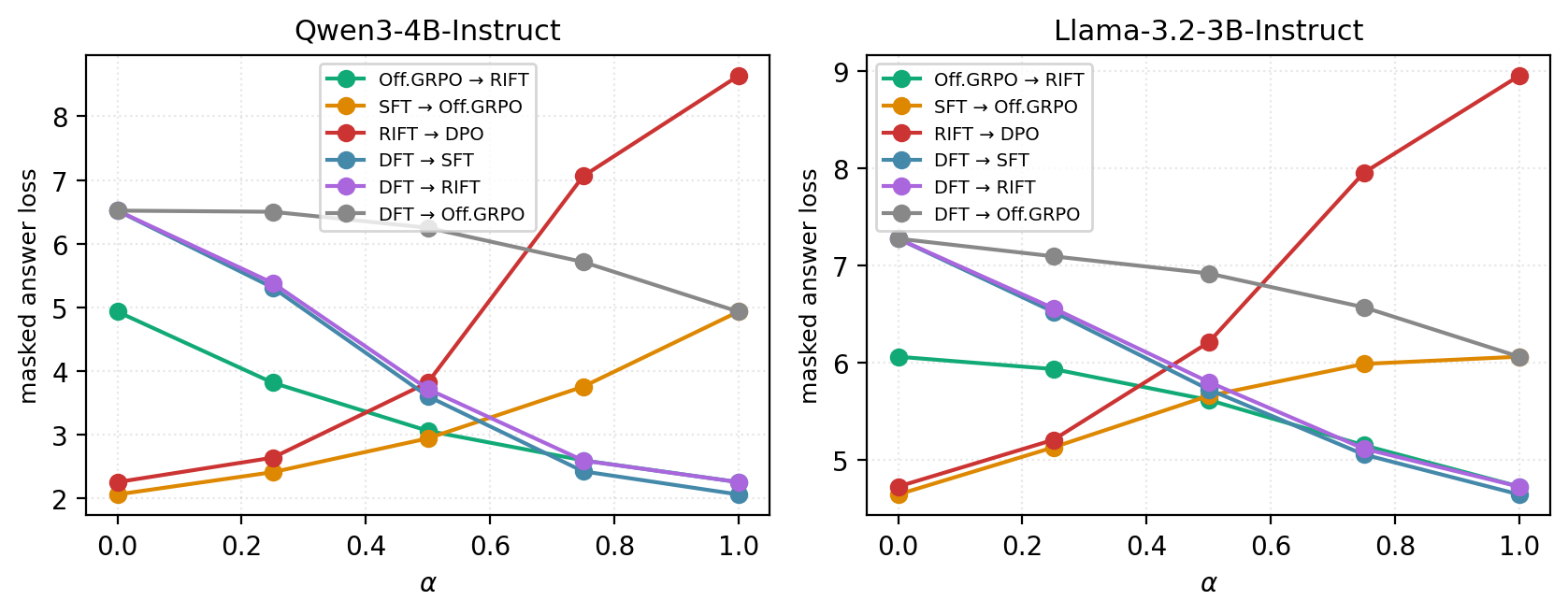}
  \caption{Linear mode connectivity (masked-answer CE on GSM8K) on Qwen3-4B (left) and Llama-3.2-3B (right). Same picture: SFT/Off.GRPO/RIFT/DFT pairs are barrier-free; RIFT$\to$DPO shows a sharp barrier above $\alpha{=}0.5$ on both architectures (DPO endpoint loss $8.64$ Qwen3, $8.96$ Llama32).}
  \label{fig:lmc}
\end{figure}

\begin{figure}[h]
  \centering
  \includegraphics[width=\linewidth]{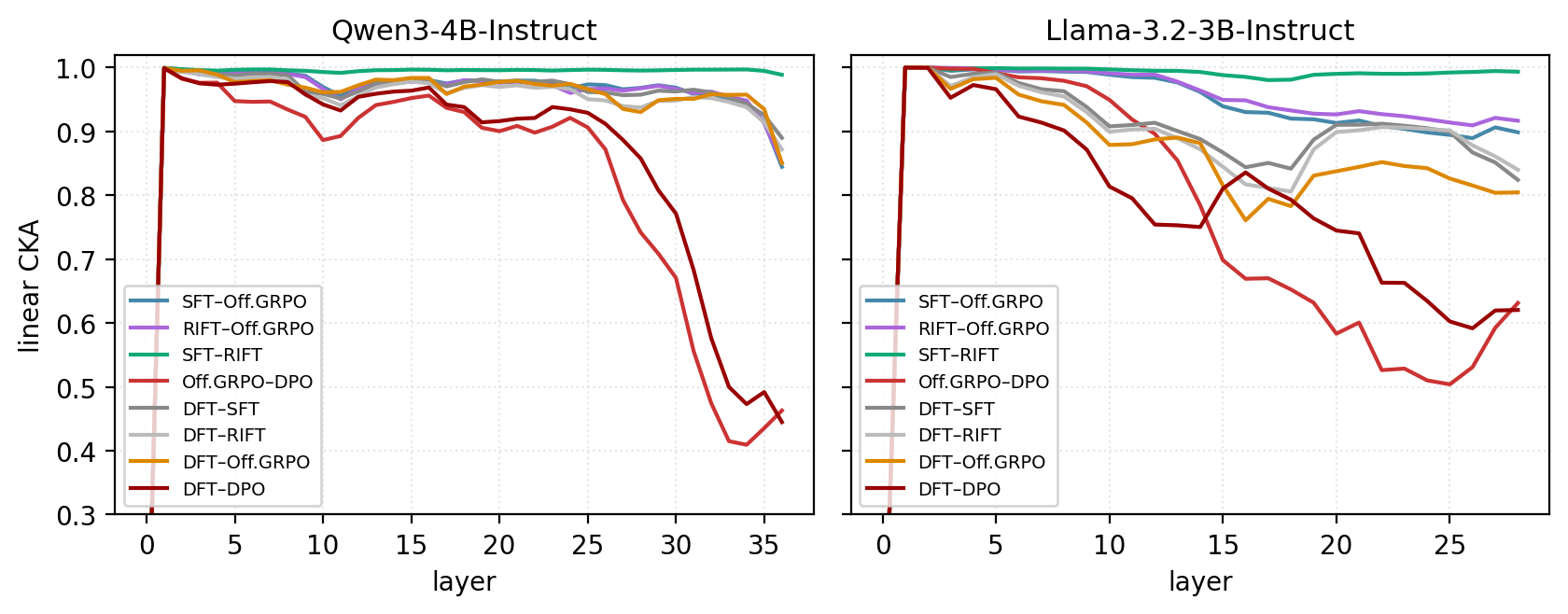}
  \caption{Linear CKA of hidden states across all blocks for selected method pairs on $100$ GSM8K prompts: Qwen3-4B (left, $36$ blocks), Llama-3.2-3B (right, $28$ blocks). On both architectures: SFT/RIFT indistinguishable ($>0.99$), Off.GRPO diverges in output-facing layers, and DPO collapses in the final third (Qwen3 $\sim\!\!0.45$, Llama32 $\sim\!\!0.62$).}
  \label{fig:cka}
\end{figure}

\end{document}